\documentclass[submission,copyright,creativecommons]{eptcs}
\providecommand{\event}{FMAS 2025} 
\usepackage{iftex}
\ifpdf
  \usepackage{underscore}         
  \usepackage[T1]{fontenc}        
\else
  \usepackage{breakurl}           
\fi
\usepackage{amssymb}
\usepackage{amsmath}
\usepackage{graphicx} 
\usepackage{enumitem}
\usepackage{amsfonts}
\usepackage{algorithm}
\usepackage{algpseudocode}
\usepackage{csquotes}

\title{Achieving Safe Control Online through Integration of Harmonic
  Control Lyapunov-Barrier Functions with Unsafe Object-Centric Action
  Policies}
\author{Marlow Fawn \qquad\qquad Matthias Scheutz
\institute{
  Human-Robot Interaction Laboratory\\Tufts University\\Medford MA 02155, USA}
\email{\quad marlow.fawn@tufts.edu \quad\qquad matthias.scheutz@tufts.edu}
}

\def\titlerunning{Online Safe Control}
\def\authorrunning{M. Fawn \& M. Scheutz}

\begin{document}
\maketitle

\begin{abstract}
  We propose a method for combining Harmonic Control Lyapunov–Barrier
  Functions (HCLBFs) derived from Signal Temporal Logic (STL)
  specifications with any given robot policy to turn an unsafe policy
  into a safe one with formal guarantees.  The two components are
  combined via HCLBF-derived safety certificates, thus producing
  commands that preserve both safety and task-driven behavior.  We
  demonstrate with a simple
  proof-of-concept implementation for an object-centric force-based
  policy trained through reinforcement learning for a movement task of
  a stationary robot arm that is able to avoid colliding with
  obstacles on a table top after combining the policy with the safety
  constraints.  The proposed method can be generalized to more complex
  specifications and dynamic task settings.
\end{abstract}

\section{Introduction}

Open-world environments pose many challenges for autonomous robots as
unexpected events or task modulations can make learned robot behavior
inapplicable or obsolete.  Consider, for example, a robot that has
learned to autonomously perform a sorting task on a table top without
any human interventions when a human co-worker steps in to help with
finishing the task.  This change in task environment now requires the
robot to avoid colliding with the human whose arms are extended into
the robot's work space and are dynamically changing position.  Even if
the robot has the perceptual capability to detect and track the
human's arms and hands, its trained action policy does not provide a
way to account for the motion constraints they impose.  Or consider a
delivery robot in a warehouse that has an optimized policy for
traversing indoor spaces when dynamic constraints are imposed on
where it can drive (e.g., because parts of the floor are painted).  In
both cases it would be important for the robot to be able to adjust
its behavior on the fly to account for the changes in the task
environment, ideally with formal guarantees that these changes are
{\em safe}.

State-of-the-art learning methods for training robots on tasks such as
Reinforcement Learning (RL) enable robots to learn complex behaviors
through trial and error, in some cases with safety guarantees
such as in the case of ``safe RL'' \cite{guetal24}, but the trained
behaviors cannot be easily modified or adapted to new circumstances or
safety constraints.  In this paper, we utilize analytical methods,
such as Control Lyapunov Functions (CLFs) and Control Barrier
Functions (CBFs), which offer formal guarantees of stability and
safety, to enable the adaptation of previously learned control
policies to changes in the task environment.  Specifically, we propose
to use {\em Signal Temporal Logic} (STL) to describe the changed task
constraints from which we then derive Harmonic Control
Lyapunov–Barrier Functions (HCLBFs) whose gradients can drive the
system toward its goals while avoiding unsafe regions.  We then
integrate the safety constraints with the robot's control policy
through a shared velocity representation that allows the HCLBF field
to modulate the learned policy without retraining.  Importantly, due
to the way the components are integrated, the resultant behavior is
provably safe.

There exists a small body of work integrating CBFs with STL. In
particular, \cite{SrinivasnaCoogan19,srinivasanlearning2020} both
provide robust STL CBF integration for agent navigation. Our primary
differences are twofold: (1) We expand on these method by integrating
Lyapunov-like functionality, further expanding the valid STL
definitions and (2) We use these CLBFs for \emph{synthesis} with CLBF
control laws, allowing for greater task flexibility through RL. With
our method, we ultimately hope to integrate control laws with temporal
logic, encompassing these two methods in a larger framework that
allows for dynamically safe control of pre-existing behaviors.

The rest of the paper is structured as follows: After briefly
reviewing some formal preliminaries, we introduce the proposed method
for deriving STL-based HCLBFs using a restricted version of STL and
Laplace solver for generating HCLBFs on a grid rather than a
continuous spatial environment.  Next, we introduce the demonstration
environment in which we train an ``object-centric force-based RL
policy'' (e.g., \cite{fangetal25icra}), i.e., a policy that represents
forces applied to object in the direction of its intended motion, in a
simple proof-of-concept arm movement task in the 2D plane from a start
to a goal position.  We then show how we can integrate both components
into a unified control loop with HCLBF certificates so that the robot
avoids colliding with unexpected obstacles.  Finally, we briefly
discuss the advantages and limitations of our proposed approach as
well as directions for future work, and conclude with a brief summary
of what we accomplished.

\section{Preliminaries}

We first briefly review three different formalisms: Signal Temporal
Logic (STL), Harmonic Control Lyapunov–Barrier Functions (HCLBFs),
and Soft Actor Critic (SAC) Reinforcement Learning (RL).

\subsection{Signal Temporal Logic (STL)}

Signal Temporal Logic (STL) extends linear temporal logic (LTL) which
is frequently used as a specification language for robot behavior
(e.g., \cite{lignosetal15}) by allowing predicates to range over real
values (e.g., spatial positions) in real-time.  I.e., we assume we are
given signals of the form $x_1(t),\ldots,x_n(t)$ which we can use in
atomic positions $p$ of the form

\[
p := f(x_1(t),\ldots,x_n(t)) \geq 0
\]  

\noindent where \( f: \mathbb{R}^n \to \mathbb{R} \) is a {\em
  constraint} (e.g., see \cite{10.1007}). Atomic
positions can be modulated with two temporal operators
\(\mathbf{G}_{[a,b]}\) and \(\mathbf{F}_{[a,b]}\) inherited from
Metric Temporal Logic (MTL), meaning ``for all times in [a,b]'' and
``there is a time in [a,b]'', respectively, thus allowing for the
specification of both temporal and spatial constraints.

The satisfaction of an STL formula $\varphi$ by a signal
$x=(x_1,\ldots,x_n)$ at time $t$ is then defined as follows:

\begin{itemize}
\item $(x,t) \models p \leftrightarrow f(x_1(t),\ldots,x_n(t)) > 0$

\item $(x,t) \models \lnot\varphi \leftrightarrow (x,t) \neg\models\varphi$

\item $(x,t) \models \varphi \land \psi \leftrightarrow (x,t) \models\varphi \land (x,t) \models \psi$

\item $(x,t) \models \varphi \land \psi \leftrightarrow (x,t) \models\varphi \land (x,t) \models \psi$

\item $(x,t)\models F[a,b]\varphi \leftrightarrow \exists t'\in[t+a,t+b]: (x,t')\models \varphi$

\end{itemize}

For our purposes here, we will focus on a small subset of STL, namely
the set of atomic formulas in the scope of one of the temporal
operators and their conjunctions (e.g., $\mathbf{G}_{[a,b]}
f(x_1(t))>0 \land \mathbf{F}_{[a',b']}f'(x_1(t),x_2(t))>0$).

\subsection{Harmonic Control Lyapunov–Barrier Functions (HCLBFs)}

{\em Control Lyapunov Barrier Functions} (CLBFs) define a framework
for combining the reachability of {\em Control Lyapunov Functions}
with the forward-invariance of {\em Control Barrier Functions} (e.g.,
\cite{ROMDLONY201639}).  Given the control function:

\[
\dot{s}(t) = f(s(t), u(t)), \quad s \in \mathbb{R}^n, \ u \in \mathbb{R}^m,
\]

\noindent a CLBF \( V: \mathcal{S} \to \mathbb{R} \) is defined as the constraints on that function such that:  
(i) \(V\) decreases along trajectories, which drives the system towards a goal and 
(ii) \(V\) has high magnitude in unsafe regions, preventing constraint violation.

We will use the definition for HCLBFs provided by Mukherjee et
al. \cite{mukherjee2024harmoniccontrollyapunovbarrier}. Let \(
\mathcal{S} \subset \mathbb{R}^n \) denote the safe set, \(
\mathcal{S}_{\text{goal}} \subset \mathcal{S} \) the goal region, and
\( \mathcal{S}_{\text{unsafe}} \subset \mathbb{R}^n \setminus
\mathcal{S} \) the unsafe set.  A function \( V \in C^2(\mathcal{S})
\cup C^1(\partial \mathcal{S}) \) is a \emph{harmonic CLBF} if it
satisfies:

\begin{enumerate}[noitemsep]
    \item \( \nabla^2 V(s) = 0 \quad \forall s \in \mathcal{S}_{\text{safe}} \);
    \item \( V(s) = 0 \quad \forall s \in \overline{\mathcal{S}_{\text{goal}}} \);
    \item \( V(s) = c \quad \forall s \in \partial \mathcal{S} \cup \mathcal{S}_{\text{unsafe}} \),
\end{enumerate}

for some constant \( c > 0 \). Intuitively, 
\begin{enumerate}[noitemsep]
\item Ensure that the field is never ``flat'' in safe regions, meaning the system will never stall
\item Ensure the goal is always the ``lowest'' point in the state space, meaning the system will be drawn to it
\item Ensure the obstacles are always the ``highest'' points in the state space, meaning the system will be repelled from them

\end{enumerate}

Formally, \(V\) is the solution to the boundary value problem defined
by Laplace's equation on \(\mathcal{S}\) with Dirichlet boundary
conditions set by the goal and unsafe sets.  In other words, we are
solving for some function \(V\) over the state space \(\mathcal{S}\)
which contains our goal, safe, and unsafe regions, such that the
position of an object following \(V\) as a trajectory will never
overlap with unsafe regions, and will ultimately be driven to goal
regions.

We will, in our proof-of-concept implementation in
Section~\ref{sec:proof-of-concept}, approximate this continuous PDE on
a finite grid for tractability.

\subsection{Reinforcement Learning}

Robot tasks and thus task-based control problems are commonly modeled
as either fully or partially observable Markov Decision Process (MDP)
\((\mathcal{S}, \mathcal{A}, P, r, \gamma)\) where \(\mathcal{S}\) is
the state space, \(\mathcal{A}\) the action space, \(P\) the
transition function, \(r: \mathcal{S} \times \mathcal{A} \to
\mathbb{R}\) the reward function, and \(\gamma \in (0,1]\) the
  discount factor.  A stochastic policy \(\pi_\theta(a \mid s)\) with
  parameters \(\theta\) induces a distribution over trajectories, and
  is optimized to maximize the expected discounted return
\[
J(\pi_\theta) = \mathbb{E}_{\tau \sim \pi_\theta} \left[ \sum_{t=0}^\infty \gamma^t r(s_t, a_t) \right].
\]

A common method for learning a policy $\pi_\theta$ is Q-learning which
learns a Q-function $Q(s,a)$ for every state $s \in \mathcal{S}$ and
action $a\in \mathcal{A}$ through (possibly unsafe) explorations of
the state space where $Q(s,a)$ is the expected discounted sum of
future rewards received by performing action $a$ in state $s$. The
Q-function is updated as follows:

\[ Q \gets Q + \alpha \left[ \left(\mathbb{R} + \gamma \max_{a^\prime
    \in \mathcal{A}} Q(a',s') - Q(a,s) \right) \right]\]

\noindent where $\alpha \in (0, 1]$ is the learning rate. Q-learning
  uses an $\epsilon$-greedy policy for exploring new states in which
  the agent selects a random action with probability $\epsilon$ rather
  than performing the action prescribed by the largest $q$-value.

The ``Soft Actor Critic'' (SAC) version of RL, consisting of an
``actor'' and a ``critic'' (often more than one critic), is an
``off-policy'' method that integrates ``maximum entropy''
reinforcement learning to maximize both the expected reward as well as
the entropy of the policy.  The actor adds an entropy term at state
$s$ to the expected discounted return:

\[
\beta\cdot -\mathbb{E}_{a} \sim \pi_\theta log \pi(a|s)
\]

\noindent where $\beta$ scales the entropy term.  The critic, in turn,
computes and tracks $Q(a,s)$.

\section{STL-Based HCLBFs}

We start by defining the permissible formal constraints in STL,
followed by a description of our HCLBF implementation. We then explain how we use STL to generate HCLBF constraints,
with a brief mention of the particular restrictions for our case based
on the resulting potential field.

\subsection{STL Definitions}

Previous work has shown that CLBFs can be used to satisfy STL formulae
for a dynamic system (e.g., \cite{9483028,8404080}). We use a
simplified version of this work to construct boundaries for HCLBFs
using a small fragment of STL which is defined as follows:

\begin{itemize}[noitemsep,leftmargin=1.3em]
    \item \textbf{Atomic formulas} $p${\bf :} We consider atomic
      formulas $p$ of the form
    \[
    x \bowtie c \quad\text{or}\quad y \bowtie c,
    \]
    \noindent where $\bowtie \in \{\ge,>,=\}$, $c$ is a
    constant, and $x$ and $y$ are components of a spatial coordinate $\langle
    x,y\rangle$.  All atomic formulas are allowable formulas.
    \item \textbf{Negation:} Only atomic formulas $p$ can be negated,
      $\lnot p$.  Negated and non-negated atomic formulas are considered
      ``allowable formulas''.
  \item \textbf{Boolean composition:} Allowable formulas $\varphi$ and
    $\psi$ can be combined through ($\wedge$), $\varphi\land\psi$; no
    other combination such as disjunction ($\vee$) or implication
    ($\Rightarrow$) is allowed.
  \item \textbf{Temporal operators:} Allowable formulas can get
    temporally quantified with a single bounded ``always'' or
    ``eventually'' operator:
    \[
    \mathbf{G}_{[t_1,t_2]} \psi \quad \text{or} \quad \mathbf{F}_{[t_1,t_2]} \psi,
    \]
    \noindent where $0 \le t_1 < t_2$; nesting of temporal operators
    is not allowed.
\end{itemize}

\noindent Examples of formulas in the restricted STL fragment are
${F}_{[t_1,t_2]}(x>0 \land \lnot x>2)$ or ${G}_{[t_1,t_2]}(x>0 \land
y>0)$.  The fragment allows for specifying intervals during which
spatial constraints need to hold, e.g., in the example from the
introduction one such specification would be to impose that the robot
not move into a spatially defined bounding box around the human's arms
{\em while} the human is operating in the robot's space, which may need to be updated dynamically.  The
specification also allows for specifying constraints that will occur
within a particular interval (e.g., that there will be a moment when
operating in a particular region is forbidden).  Note that we
interpret constants here as making spatial references to discrete grid
coordinates in the robot's working space, but nothing critical hinges
on it (we will later extend this to continuous domains through
interpolation outside STL).

\subsection{HCLBF Definitions}

To reiterate from our section 2.2, in theory, HCLBFs are defined as scalar functions that solve a
boundary value problem on a continuous safe set. Formally, we define
\[
V : \mathbb{R}^2 \to [0,1],
\]
where the associated velocity field is obtained by the negative gradient
\[
U_\Phi(x,y) = -\nabla V(x,y),
\]

\noindent which provides directions of motion that flow toward goals
while avoiding unsafe regions. Intuitively, an HCLBF is a single function that defines a stable policy for how to move across a space such that certain constraints will be followed with formal guarantees.

In our implementation, we discretize the HCLBF
onto a finite grid for tractability and to align with our discrete STL
construction of spatial restrictions. The potential field $V$ is thus
stored at integer grid points $(i,j) \in \mathbb{Z}^2$, and gradients
are approximated using finite differences.

Similar to our STL restrictions, we impose some limitations on the
form of the HCLBF that can be relaxed in future work:

\begin{itemize}
    \item In formal HCLBFs, goal and unsafe regions cannot overlap. However, in real-world scenarios this happens frequently (e.g. an obstacle is placed or moves over a known goal region). As our goal is to prioritize safety, we allow the unsafe regions to overrule goal regions. While this leads to unreachable scenarios, it ensures that all scenarios are fully safe.
    \item We ignore settings where the agent is already in collision with an obstacle. This is solvable with heuristics driving the agent away from the obstacle region, but is irrelevant for our intended use of fully safe control.
    \item At least one goal cell is required, otherwise the velocity
      field degenerates to trivial flow around unsafe regions.
    \item The current formulation is grid-based, and gradients are
      approximated numerically as continuous extensions would require
      analytic PDE solutions or finer interpolation, which is
      computationally expensive.
\end{itemize}

\begin{figure}
  \centering
  \includegraphics[width=.45\linewidth]{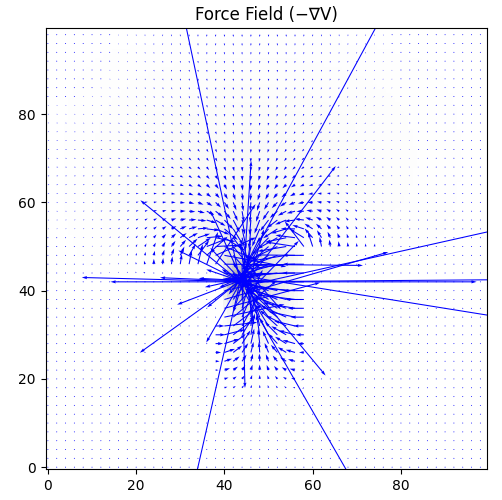}
  \includegraphics[width=.45\linewidth]{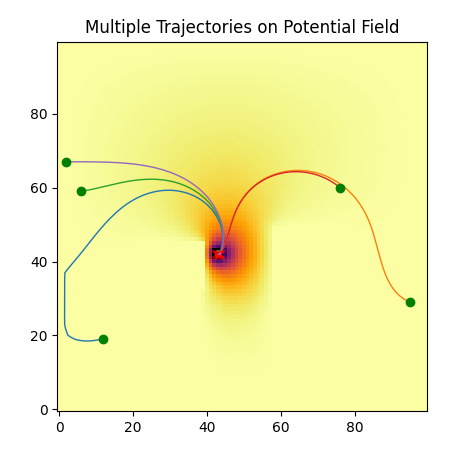}
  \caption{HCLBF field and gradient (left) and sample trajectories
    demonstrating safe goal reaching (right).}
  \label{fig:hclbf}
\end{figure}

In sum, the use of HCLBF with STL is best understood as a
potential field encoding of the STL specification: It takes in a
spatial state $(x,y)$ and outputs a velocity vector $U_\Phi(x,y)$ that
balances progress toward the goal against avoidance of unsafe regions
(see Fig.~\ref{fig:hclbf} for examples).

\subsection{Laplace Solution for HCLBFs}

To generate an HCLBF from a given STL formula, we translate the
logical specification into spatial boundary conditions and then solve
a Laplace equation on the workspace grid.  The STL construction
produces two sets, a goal region $G$ and an unsafe region $U$, which
can be converted into Dirichlet boundary conditions for the scalar
potential $V$:
\[
V(x,y) = 0 \quad \text{for } (x,y)\in G, 
\qquad
V(x,y) = 1 \quad \text{for } (x,y)\in U.
\]

On the remaining safe cells, we enforce harmonicity:
\[
\nabla^2 V(x,y) = 0,
\]

\noindent which yields a unique smooth interpolation between goal and
unsafe regions. In practice, $V$ is computed on the discrete grid
using an iterative relaxation method until convergence.

\section{Proof-of-concept Setting}
\label{sec:proof-of-concept}

To demonstrate how the above method works to safeguard behavior of
autonomous robots when there are changes to a task environment, we
consider a simple robot arm navigation task.  In this task, the robot
arm has to traverse the surface of table top without any obstacles
moving from an initial position to a goal position.  We start by
training an RL policy, as would be commonly done for robot tasks
(usually more complex ones, of course) using an object-centric
force-based policy (e.g., \cite{fangetal25icra}).

\subsection{Policy Learning}

We train the object-centric policy with a Soft Actor–Critic (SAC) RL
algorithm on a rigid ``virtual cube'', i.e., a proxy for the motion of
the robot, learning how to move towards randomly sampled goals in task
space.  Since training occurs in \emph{object mode}, i.e., forces
applied directly to the cube, we can keep the policy independent of
the particular robot and robot morphology and thus allow for the
trained policy to operate on any robot for which an inverse kinematics
model exists (e.g., see \cite{fangetal25icra} for details on the
method and a discussion).  We will then later use the same learned
representations when deploying the policy through an ``Operational
Space Control'' (OSC) \cite{1087068} interface.

Note that the policy observes the object's Cartesian position and
velocity,
\[
s_t = [x(t), v(t)] \in \mathbb{R}^{2n},
\]

\noindent and outputs a task-space force \(a_t = F_t \in
\mathbb{R}^n\) (in our experiments \(n=2\) for planar
motion).  We use a dense shaping signal
\[
r_t \;=\; -\|x(t) - x_{\mathrm{goal}}\|_2 \;-\; 0.01,
\]
\noindent with a success bonus and out-of-bounds penalty.

In sum, using force as the action space achieves three goals: (i) it
keeps the policy object-centric and robot-agnostic, (ii) it produces
commands in a common kinematic space that can be directly used with
HCLBFs, and (iii) it leverages existing OSC infrastructure for
low-level actuation without requiring direct torque control.

\subsection{Robot Controller}

We assume a rigid, no-slip grasp between the robot's end effector (EE)
and the controlled object, so that the object twist equals the EE
twist, \(V_{\text{obj}} = V_{\text{ee}}\).  This allows forces applied
to the object in training to be mapped directly and consistently to EE
commands on the robot.

We convert the force \(F_d\) outputs of learned policy (on the cube)
to a desired EE velocity \(U_\pi(x)\) using an admittance model $A(x)$
(e.g., \cite{4788393}), so that commanded forces translate to
proportional velocities in task space.  For our implementation, the
simplified admittance model where the force $F_d$ is scaled by a
constant gain to produce a velocity $U_\pi(x) = \alpha F_d$ is
sufficient but nothing hinges on it and more complex models are
possible.  This model preserves the direction encoded by the policy
while providing a tuning for integration into the OSC. It also allows
for future substitutions with more complex or kinematically accurate
models without affecting the safety of the system.

Since the robot's OSC accepts Cartesian pose targets, we integrate the
commanded velocity over the control period \(\Delta t\) to update the
EE position:

\[
p_{t+1} = p_t + U_\pi(x_t)\, \Delta t,
\]

\noindent with orientation held fixed.

\section{Control Synthesis}

We now demonstrate how we can combine the HCLBF safety constraints
with the learned RL policy.  And note that while we are using RL here,
as it is commonly used in robotics, in principle any OSC controller
could be used in its place (e.g., online motion planners); These controllers would
maintain the safety guarantees provided by the HCLBF filtering regardless of underlying controller.

Let
\[
u = U_\pi(x) := \mathcal{A}(x)\,F_\pi(x)
\]

\noindent denote the nominal velocity command obtained from the policy
via the admittance map \(\mathcal{A}(x)\). The HCLBF scalar function
\(V(x)\) and gradient \(\nabla V(x)\) are evaluated by mapping the
current world position to grid indices and interpolating both the
potential and its finite-difference derivatives, converted to world
coordinates.

The safety condition is expressed as the barrier inequality
\[
\nabla V(x)^\top \tilde u \;\le\; -k_\alpha V(x),
\]

\noindent where \(k_\alpha > 0\) is a constant gain.  
If the nominal \(u\) satisfies this inequality, it is accepted. Otherwise, we solve the projection problem
\[
u^\star = \arg\min_{\tilde u \in \mathbb{R}^2} \;\|\tilde u - u\|_2^2 
\quad \text{s.t.} \quad \nabla V(x)^\top \tilde u \le -k_\alpha V(x).
\]

\noindent This ensures the final action is the closest possible safe
modification of the policy output.

The controller thus applies the HCLBF safety filter to minimally
correct the learned policy's action as follows:

\begin{enumerate}[noitemsep]
    \item Query the learned policy to obtain a task-space force \(F_\pi(x)\).
    \item Map the force to a task-space velocity \(u = U_\pi(x)\) using an admittance model.
    \item Interpolate the HCLBF scalar field \(V(x)\) and its gradient \(\nabla V(x)\) at the current end-effector position.
    \item Check the barrier inequality \(\nabla V(x)^\top u \le -k_\alpha V(x)\). If satisfied, keep \(u\). If violated, project \(u\) onto the safe half-space to obtain the corrected velocity \(u^\star\).
    \item Package the final command with vertical and gripper components, integrate over the control period, and send the pose target to the robot’s OSC controller.
\end{enumerate}

Note that the above algorithm ensures that safety constraints are
enforced at every step without retraining or modifying the learned
policy.

We then extend corrected velocity \(u^\star\) with vertical and gripper components to form
\[
u_{\text{cmd}} = [u^\star_x,\; u^\star_y,\; z_{\text{hold}}-z_{\text{ee}},\; g_{\text{default}}],
\]
which is integrated over the control period \(\Delta t\) to update the end-effector target:
\[
p_{t+1} = p_t + u^\star\,\Delta t.
\]

\begin{figure}
  \centering
    \includegraphics[width=.8\linewidth]{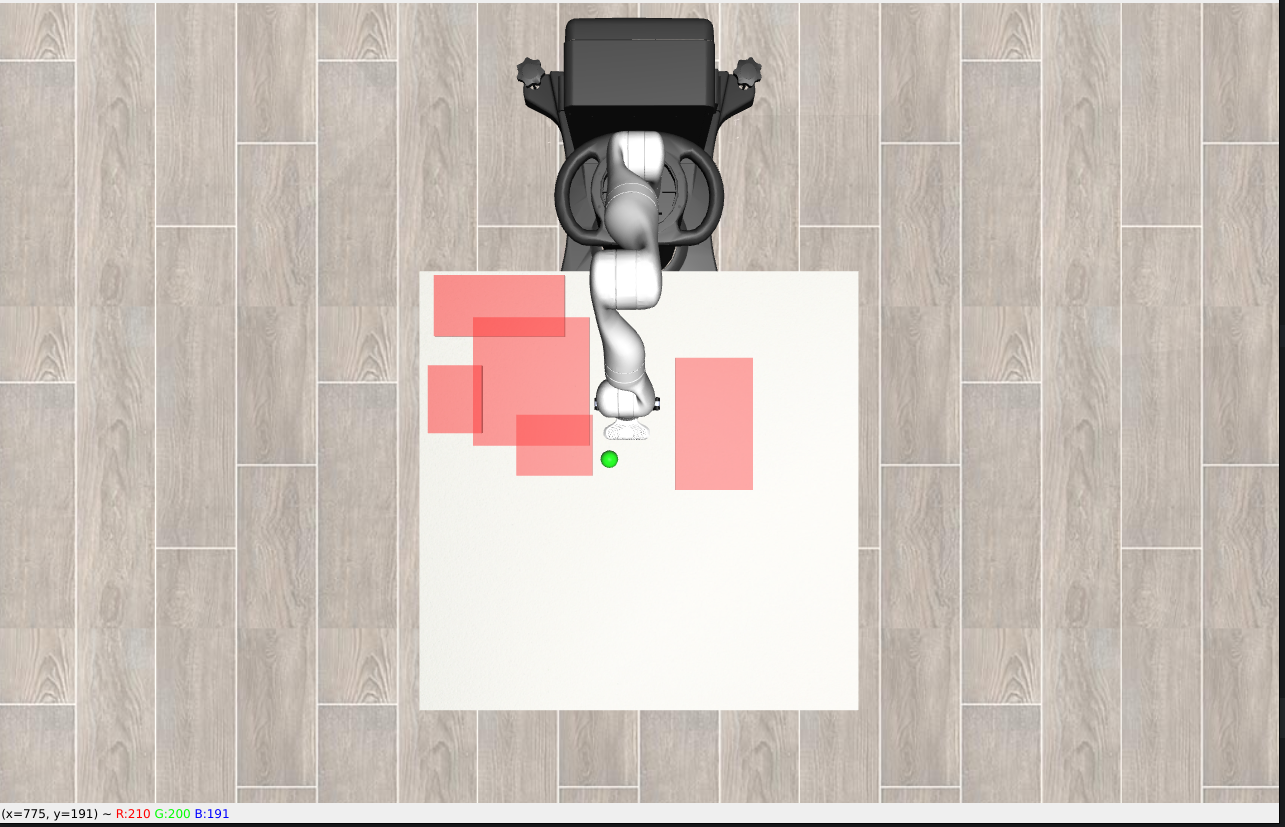}
    \caption{An image from the implementation of the above method in
      the {\em RoboSuite} simulation environment showing the task
      arena with obstacles (red) and the robot arm's goal point
      (green).}
    \label{fig:env}
\end{figure}

\noindent The new position \(p_{t+1}\) is passed to the
\texttt{OSC\_POSITION} controller, which computes joint commands.  The
environment (simulation or hardware) then advances one step, and the
cycle repeats until termination.  Fig.~\ref{fig:env} shows the
experimental evaluation setting in the {\em RoboSuite} robotics
simulation environment.

\section{Discussion}

The proof-of-concept demonstration shows the viability of the proposed
method for generating safety guardrails for unsafe policies online.
However, the online computation speed will depend on multiple
factors, including the complexity of the safety specifications, both
in terms of the spatial regions but also the temporal complexity and
duration of the intervals as well as the granularity of the
discretization and convergence tolerance of the solver. For
constraints with very short intervals arriving in real-time, the time
to calculate the safety velocity field might exceed the constraint
time.  Also, extensions to higher-dimensional workspaces will require
efficient solvers which can generate sufficient approximations
quickly.  On the other hand, if the setting is static or does not
change frequently, there will be sufficient time to calculate the
potentials and the computational effort and time to do so will likely
be significantly shorter than adapting the robot's control through RL
or training it from scratch.

It is also interesting to consider the proposed approach in light of
work in safe RL (e.g., \cite{guetal24}) where a ``safety shield'' is
provided initially that bounds the region within which the robot can
explore and learn its policy.  The current approach could be used in a
similar way during regular RL training when the robot needs to learn
an action policy as the ``exploration'' part of RL might make the
robot attempt actions that lead to unsafe regions -- using the safety
HCLBF the robot would be prevented from carrying out unsafe actions.

The current method is not without several limitations that will be
addressed in future work.  For one, we used a very limited subset of
STL for specifying only spatial constraints in order to simplify the
construction of HCLBFs. Extending the usable subset of STL to richer
logical structures, including disjunctions and nested temporal operators,
would allow for more complex specifications of time-varying goals and
constraints.  It would also be important to explore the use of
continuous PDE solvers to be able to replace the current grid-based
Laplace approximation, which would enable higher-dimensional
workspaces and real-time updates. Existing approaches to explore include sum-of-squares methods \cite{dai2024verificationsynthesiscompatiblecontrol} and neural CBFs \cite{zhang2024seevsynthesisefficientexact}. Finally, the current implementation follows theory to create fully safe navigation, and initial trials have proven successful. However, this must be evaluated formally with quantitative analyses to demonstrate safety in regular operation.

\section{Conclusion}

In this paper we proposed a new way of adapting an existing robot
policy (e.g., trained through reinforcement learning) to meet new
(potentially dynamic) motion constraints without requiring retraining
of the policy.  The proposed method utilizes Harmonic Control
Lyapunov–Barrier Functions generated from formal constraints in Signal
Temporal Logic which are then combined with the learned policy in a
shared velocity space.  The method enables online adaptations
of robot behavior that inherit the formal guarantees of Harmonic
Control Lyapunov–Barrier Functions, providing a unified control
framework that preserves both task performance and safety.


\bibliographystyle{eptcs}
\bibliography{references.bib}

\begin{thebibliography}{10}
\providecommand{\bibitemdeclare}[2]{}
\providecommand{\surnamestart}{}
\providecommand{\surnameend}{}
\providecommand{\urlprefix}{Available at }
\providecommand{\url}[1]{\texttt{#1}}
\providecommand{\href}[2]{\texttt{#2}}
\providecommand{\urlalt}[2]{\href{#1}{#2}}
\providecommand{\doi}[1]{doi:\urlalt{https://doi.org/#1}{#1}}
\providecommand{\eprint}[1]{arXiv:\urlalt{https://arxiv.org/abs/#1}{#1}}
\providecommand{\bibinfo}[2]{#2}

\bibitemdeclare{misc}{dai2024verificationsynthesiscompatiblecontrol}
\bibitem{dai2024verificationsynthesiscompatiblecontrol}
\bibinfo{author}{Hongkai \surnamestart Dai\surnameend},
  \bibinfo{author}{Chuanrui \surnamestart Jiang\surnameend},
  \bibinfo{author}{Hongchao \surnamestart Zhang\surnameend} \&
  \bibinfo{author}{Andrew \surnamestart Clark\surnameend}
  (\bibinfo{year}{2024}): \emph{\bibinfo{title}{Verification and Synthesis of
  Compatible Control Lyapunov and Control Barrier Functions}},
  \doi{10.48550/arXiv.2406.18914}.
\newblock \urlprefix\url{https://arxiv.org/abs/2406.18914}.

\bibitemdeclare{inproceedings}{10.1007}
\bibitem{10.1007}
\bibinfo{author}{Alexandre \surnamestart Donz{\'e}\surnameend}
  (\bibinfo{year}{2013}): \emph{\bibinfo{title}{On Signal Temporal Logic}}.
\newblock In \bibinfo{editor}{Axel \surnamestart Legay\surnameend} \&
  \bibinfo{editor}{Saddek \surnamestart Bensalem\surnameend}, editors:
  {\slshape \bibinfo{booktitle}{Runtime Verification}},
  \bibinfo{publisher}{Springer Berlin Heidelberg}, \bibinfo{address}{Berlin,
  Heidelberg}, pp. \bibinfo{pages}{382--383},
  \doi{10.1007/978-3-642-40787-1_27}.

\bibitemdeclare{inproceedings}{fangetal25icra}
\bibitem{fangetal25icra}
\bibinfo{author}{Shijie \surnamestart Fang\surnameend},
  \bibinfo{author}{Wenchang \surnamestart Gao\surnameend},
  \bibinfo{author}{Shivam \surnamestart Goel\surnameend},
  \bibinfo{author}{Christopher \surnamestart Thierauf\surnameend},
  \bibinfo{author}{Matthias \surnamestart Scheutz\surnameend} \&
  \bibinfo{author}{Jivko \surnamestart Sinapov\surnameend}
  (\bibinfo{year}{2025}): \emph{\bibinfo{title}{{FLEX}: A Framework for
  Learning Robot-Agnostic Force-based Skills Involving Sustained Contact Object
  Manipulation}}.
\newblock In: {\slshape \bibinfo{booktitle}{Proceedings of the 2025 IEEE
  International Conference on Robotics and Automation}}, pp.
  \bibinfo{pages}{4782--4788}, \doi{10.1109/ICRA55743.2025.11127866}.

\bibitemdeclare{article}{guetal24}
\bibitem{guetal24}
\bibinfo{author}{Shangding \surnamestart Gu\surnameend}, \bibinfo{author}{Long
  \surnamestart Yang\surnameend}, \bibinfo{author}{Yali \surnamestart
  Du\surnameend}, \bibinfo{author}{Guang \surnamestart Chen\surnameend},
  \bibinfo{author}{Florian \surnamestart Walter\surnameend},
  \bibinfo{author}{Jun \surnamestart Wang\surnameend} \& \bibinfo{author}{Alois
  \surnamestart Knoll\surnameend} (\bibinfo{year}{2024}):
  \emph{\bibinfo{title}{A Review of Safe Reinforcement Learning: Methods,
  Theories, and Applications}}.
\newblock {\slshape \bibinfo{journal}{IEEE Transactions on Pattern Analysis and
  Machine Intelligence}} \bibinfo{volume}{46}(\bibinfo{number}{12}), pp.
  \bibinfo{pages}{11216--11235}, \doi{10.1109/TPAMI.2024.3457538}.

\bibitemdeclare{inproceedings}{4788393}
\bibitem{4788393}
\bibinfo{author}{Neville \surnamestart Hogan\surnameend}
  (\bibinfo{year}{1984}): \emph{\bibinfo{title}{Impedance Control: An Approach
  to Manipulation}}.
\newblock In: {\slshape \bibinfo{booktitle}{1984 American Control Conference}},
  pp. \bibinfo{pages}{304--313}, \doi{10.23919/ACC.1984.4788393}.

\bibitemdeclare{article}{1087068}
\bibitem{1087068}
\bibinfo{author}{O.~\surnamestart Khatib\surnameend} (\bibinfo{year}{1987}):
  \emph{\bibinfo{title}{A unified approach for motion and force control of
  robot manipulators: The operational space formulation}}.
\newblock {\slshape \bibinfo{journal}{IEEE Journal on Robotics and Automation}}
  \bibinfo{volume}{3}(\bibinfo{number}{1}), pp. \bibinfo{pages}{43--53},
  \doi{10.1109/JRA.1987.1087068}.

\bibitemdeclare{article}{lignosetal15}
\bibitem{lignosetal15}
\bibinfo{author}{Constantine \surnamestart Lignos\surnameend},
  \bibinfo{author}{Vasumathi \surnamestart Raman\surnameend},
  \bibinfo{author}{Cameron \surnamestart Finucane\surnameend},
  \bibinfo{author}{Mitchell~P. \surnamestart Marcus\surnameend} \&
  \bibinfo{author}{Hadas \surnamestart Kress{-}Gazit\surnameend}
  (\bibinfo{year}{2015}): \emph{\bibinfo{title}{Provably correct reactive
  control from natural language}}.
\newblock {\slshape \bibinfo{journal}{Auton. Robots}}
  \bibinfo{volume}{38}(\bibinfo{number}{1}), pp. \bibinfo{pages}{89--105},
  \doi{10.1007/S10514-014-9418-8}.

\bibitemdeclare{article}{8404080}
\bibitem{8404080}
\bibinfo{author}{Lars \surnamestart Lindemann\surnameend} \&
  \bibinfo{author}{Dimos~V. \surnamestart Dimarogonas\surnameend}
  (\bibinfo{year}{2019}): \emph{\bibinfo{title}{Control Barrier Functions for
  Signal Temporal Logic Tasks}}.
\newblock {\slshape \bibinfo{journal}{IEEE Control Systems Letters}}
  \bibinfo{volume}{3}(\bibinfo{number}{1}), pp. \bibinfo{pages}{96--101},
  \doi{10.1109/LCSYS.2018.2853182}.

\bibitemdeclare{misc}{mukherjee2024harmoniccontrollyapunovbarrier}
\bibitem{mukherjee2024harmoniccontrollyapunovbarrier}
\bibinfo{author}{Amartya \surnamestart Mukherjee\surnameend},
  \bibinfo{author}{Ruikun \surnamestart Zhou\surnameend},
  \bibinfo{author}{Haocheng \surnamestart Chang\surnameend} \&
  \bibinfo{author}{Jun \surnamestart Liu\surnameend} (\bibinfo{year}{2024}):
  \emph{\bibinfo{title}{Harmonic Control Lyapunov Barrier Functions for
  Constrained Optimal Control with Reach-Avoid Specifications}},
  \doi{10.48550/arXiv.2310.02869}.

\bibitemdeclare{article}{ROMDLONY201639}
\bibitem{ROMDLONY201639}
\bibinfo{author}{Muhammad~Zakiyullah \surnamestart Romdlony\surnameend} \&
  \bibinfo{author}{Bayu \surnamestart Jayawardhana\surnameend}
  (\bibinfo{year}{2016}): \emph{\bibinfo{title}{Stabilization with guaranteed
  safety using Control Lyapunov–Barrier Function}}.
\newblock {\slshape \bibinfo{journal}{Automatica}} \bibinfo{volume}{66}, pp.
  \bibinfo{pages}{39--47}, \doi{10.1016/j.automatica.2015.12.011}.
\newblock
  \urlprefix\url{https://www.sciencedirect.com/science/article/pii/S0005109815005439}.

\bibitemdeclare{misc}{srinivasanlearning2020}
\bibitem{srinivasanlearning2020}
\bibinfo{author}{Krishnan \surnamestart Srinivasan\surnameend},
  \bibinfo{author}{Benjamin \surnamestart Eysenbach\surnameend},
  \bibinfo{author}{Sehoon \surnamestart Ha\surnameend}, \bibinfo{author}{Jie
  \surnamestart Tan\surnameend} \& \bibinfo{author}{Chelsea \surnamestart
  Finn\surnameend} (\bibinfo{year}{2020}): \emph{\bibinfo{title}{Learning to be
  {Safe}: {Deep} {RL} with a {Safety} {Critic}}},
  \doi{10.48550/arXiv.2010.14603}.
\newblock \urlprefix\url{http://arxiv.org/abs/2010.14603}.

\bibitemdeclare{article}{SrinivasnaCoogan19}
\bibitem{SrinivasnaCoogan19}
\bibinfo{author}{Mohit \surnamestart Srinivasan\surnameend} \&
  \bibinfo{author}{Samuel \surnamestart Coogan\surnameend}
  (\bibinfo{year}{2019}): \emph{\bibinfo{title}{Control Of Mobile Robots Using
  Barrier Functions Under Temporal Logic Specifications}}.
\newblock {\slshape \bibinfo{journal}{CoRR}} \bibinfo{volume}{abs/1908.04903},
  \doi{10.48550/arXiv.1908.04903}.
\newblock \urlprefix\url{http://arxiv.org/abs/1908.04903}.

\bibitemdeclare{inproceedings}{9483028}
\bibitem{9483028}
\bibinfo{author}{Wei \surnamestart Xiao\surnameend}, \bibinfo{author}{Calin~A.
  \surnamestart Belta\surnameend} \& \bibinfo{author}{Christos~G. \surnamestart
  Cassandras\surnameend} (\bibinfo{year}{2021}): \emph{\bibinfo{title}{High
  Order Control Lyapunov-Barrier Functions for Temporal Logic Specifications}}.
\newblock In: {\slshape \bibinfo{booktitle}{2021 American Control Conference
  (ACC)}}, pp. \bibinfo{pages}{4886--4891},
  \doi{10.23919/ACC50511.2021.9483028}.

\bibitemdeclare{misc}{zhang2024seevsynthesisefficientexact}
\bibitem{zhang2024seevsynthesisefficientexact}
\bibinfo{author}{Hongchao \surnamestart Zhang\surnameend},
  \bibinfo{author}{Zhizhen \surnamestart Qin\surnameend},
  \bibinfo{author}{Sicun \surnamestart Gao\surnameend} \&
  \bibinfo{author}{Andrew \surnamestart Clark\surnameend}
  (\bibinfo{year}{2024}): \emph{\bibinfo{title}{SEEV: Synthesis with Efficient
  Exact Verification for ReLU Neural Barrier Functions}},
  \doi{10.48550/arXiv.2410.20326}.
\newblock \urlprefix\url{https://arxiv.org/abs/2410.20326}.

\end{thebibliography}
\end{document}